# Anytime Planning for Decentralized POMDPs using Expectation Maximization

**Akshat Kumar**
Computer Science Dept.
University of Massachusetts, Amherst
akshat@cs.umass.edu

**Shlomo Zilberstein**
Computer Science Dept.
University of Massachusetts, Amherst
shlomo@cs.umass.edu

## Abstract

Decentralized POMDPs provide an expressive framework for multi-agent sequential decision making. While finite-horizon DEC-POMDPs have enjoyed significant success, progress remains slow for the infinite-horizon case mainly due to the inherent complexity of optimizing stochastic controllers representing agent policies. We present a promising new class of algorithms for the infinite-horizon case, which recasts the optimization problem as inference in a mixture of DBNs. An attractive feature of this approach is the straightforward adoption of existing inference techniques in DBNs for solving DEC-POMDPs and supporting richer representations such as factored or continuous states and actions. We also derive the Expectation Maximization (EM) algorithm to optimize the joint policy represented as DBNs. Experiments on benchmark domains show that EM compares favorably against the state-of-the-art solvers.

## 1 Introduction

Decentralized partially observable MDPs (DEC-POMDPs) have emerged in recent years as an important framework for modeling sequential decision making by a team of agents [5]. Their expressive power makes it possible to tackle coordination problems in which agents must act based on different partial information about the environment and about each other to maximize a global reward function. Applications of DEC-POMDPs include coordinating the operation of planetary exploration rovers [3], coordinating firefighting robots [14], broadcast channel protocols [5] and target tracking by a team of sensor agents [13]. However, the rich model comes with a price–optimally solving a finite-horizon DEC-POMDP is NEXP-Complete [5]. In contrast, finite-horizon POMDPs are PSPACE-complete [12], a strictly lower complexity class that highlights the difficulty of solving DEC-POMDPs.

Recently, a multitude of point-based approximate algorithms have been proposed for solving finite-horizon DEC-POMDPs [11, 8, 16]. However, unlike their point-based counterparts in POMDPs ([15, 17]), they cannot be easily adopted for the infinite-horizon case due to a variety of reasons. For example, POMDP algorithms represent the policy compactly as $\alpha$-vectors, whereas all DEC-POMDP algorithms explicitly store the policy as a mapping from observation sequences to actions, making them unsuitable for the infinite-horizon case. In POMDPs, the Bellman equation forms the basis of most point-based solvers, but as Bernstein *et. al.* [4] highlight, no analogous equation exists for DEC-POMDPs.

To alleviate such problems, most infinite-horizon algorithms represent agent policies as finite-state controllers [1, 4]. So far, only two algorithms have shown promise for effectively solving infinite-horizon DEC-POMDPs–decentralized bounded policy iteration (DEC-BPI) [4] and a non-linear programming based approach (NLP) [1]. However, both of these algorithms have significant drawbacks in terms of the representative class of problems that can be handled. For example, solving DEC-POMDPs with continuous state or action spaces is not supported by either of these approaches. Scaling up to structured representations such as factored or hierarchical state-space is difficult due to convergence issues in DEC-BPI and a potential increase in the number of non-linear constraints in the NLP solver. Further, none of the above approaches have been shown to work for more than 2 agents, a significant bottleneck for solving practical problems.

To address these shortcomings, we present a promising new class of algorithms which amalgamates planning with probabilistic inference and opens the door to the application of rich inference techniques to solving infinite-horizon DEC-POMDPs. Our technique is based on Toussaint *et. al.*'s approach of transforming the planning problem to its equivalent mixture of dynamic Bayes nets (DBNs) and using likelihood maximization in this framework to optimize the policy value [20, 19]. Earlier work on planning by probabilistic inference can be found in [2]. Such approaches have been successful in solving MDPs and POMDPs [19]. They also easily extend to factored or hierarchical structures [18] and can handle continuous action and state spaces thanks to advanced probabilistic inference techniques [10]. We show how DEC-POMDPs, which are much harder to solve than MDPs or POMDPs, can also be reformulated as a mixture of DBNs. We then present the Expectation Maximization algorithm (EM) to maximize the reward likelihood in this framework. The EM algorithm naturally has the desirable anytime property as it is guaranteed to improve the likelihood (and hence the policy value) with each iteration. We also discuss its extension to large multi-agent systems. Our experiments on benchmark domains show that EM compares favorably against the state-of-the-art algorithms, DEC-BPI and NLP-based optimization. It always produces better quality policies than DEC-BPI and for some instances, it nearly doubles the solution quality of the NLP solver. Finally, we discuss potential pitfalls, which are inherent in the EM based approach.

## 2 The DEC-POMDP model

In this section, we introduce the DEC-POMDP model for two agents [5]. Note that finite-horizon DEC-POMDPs are NEXP complete even for two agents.

The set $S$ denotes the set of environment states, with a given initial state distribution $b_0$. The action set of agent 1 is denoted by $A$ and agent 2 by $B$. The state transition probability $P(s'|s,a,b)$ depends upon the actions of both the agents. Upon taking the joint action $\langle a,b \rangle$ in state $s$, agents receive the joint reward $R(s,a,b)$. $Y$ is the finite set of observations for agent 1 and $Z$ for agent 2. $O(s,ab,yz)$ denotes the probability $P(y,z|s,a,b)$ of agent 1 observing $y \in Y$ and agent 2 observing $z \in Z$ when the joint action $\langle a,b \rangle$ was taken and resulted in state $s$.

To highlight the differences between a single agent POMDP and a DEC-POMDP, we note that in a POMDP an agent can maintain a belief over the environment state. However, in a DEC-POMDP, an agent is not only uncertain about the environment states but also about the actions and observations of the other agent. Therefore in a DEC-POMDP a belief over the states cannot be maintained during execution time.

This added uncertainty about other agents in the system make DEC-POMDPs NEXP complete [5].

We are concerned with solving infinite-horizon DEC-POMDPs with a discount factor $\gamma$. We represent the stationary policy of each agent using a fixed size, stochastic finite-state controller (FSC) similar to [1]. An FSC can be described by a tuple $\langle N, \pi, \lambda, \nu \rangle$. $N$ denotes a finite set of controller nodes $n$; $\pi : N \to \Delta A$ represents the actions selection model or the probability $\pi_{an} = P(a|n)$; $\lambda : N \times Y \to \Delta N$ represents the node transition model or the probability $\lambda_{n'ny} = P(n'|n,y)$; $\nu : N \to \Delta N$ represents the initial node distribution $\nu_n = P(n)$. We adopt the convention that nodes of agent 1's controller are denoted by $p$ and agent 2's by $q$. Other problem parameters such as observation function $P(y,z|s,a,b)$ are represented using subscripts as $P_{yzsab}$. The value for starting the controllers in nodes $\langle p,q \rangle$ at state $s$ is given by:

$$V(p,q,s) = \sum_{a,b} \pi_{ap}\pi_{bq}\Big[R_{sab}+$$
$$\gamma \sum_{s'} P_{s'sab} \sum_{y,z} P_{yzs'ab} \sum_{p',q'} \lambda_{p'py}\lambda_{q'qz}V(p',q',s')\Big].$$

The goal is to set the parameters $\langle \pi, \lambda, \nu \rangle$ of the agents' controllers (of some given size) that maximize the expected discounted reward for the initial belief $b_0$:

$$V(b_0) = \sum_{p,q,s} \nu_p \nu_q b_0(s) V(p,q,s)$$

## 3 DEC-POMDPs as mixture of DBNs

In this section, we describe how DEC-POMDPs can be reformulated as a mixture of DBNs such that maximizing the reward likelihood (to be defined later) in this framework is equivalent to optimizing the joint policy. Our approach is based on the framework proposed in [19, 20] to solve Markovian planning problems using probabilistic inference. First we informally describe the intuition behind this reformulation (for details please refer to [19]) and then we describe in detail the steps specific to DEC-POMDPs.

A DEC-POMDP can be described using a single DBN where the reward is emitted at each time step. However, in our approach, it is described by an infinite mixture of a special type of DBNs where reward is emitted *only at the end*. For example, Fig. 1(a) describes the DBN for time $t = 0$. The key intuition is that for the reward emitted at any time step $T$, we have a separate DBN with the general structure as in Fig. 1(b). Further, to simulate the discounting of rewards, probability of time variable $T$ is set as $P(T=t) = \gamma^t(1-\gamma)$. This ensures that $\sum_{t=0}^{\infty} p_t = 1$. In addition, the random variable $r$ shown in Fig. 1(a,b) is a binary variable with its conditional distribution (for any time $T$) described using the normalized immediate reward as $\hat{R}_{sab} = P(r=1|s_T=s, a_T=a, b_T=b) = (R_{sab} - R_{min})/(R_{max} - R_{min})$. This scaling of the reward is the key to transforming the optimization problem from the realm of planning to likelihood maximization as stated below. $\theta$ denotes the parameters $\langle \pi, \lambda, \nu \rangle$ for each agent's controller.

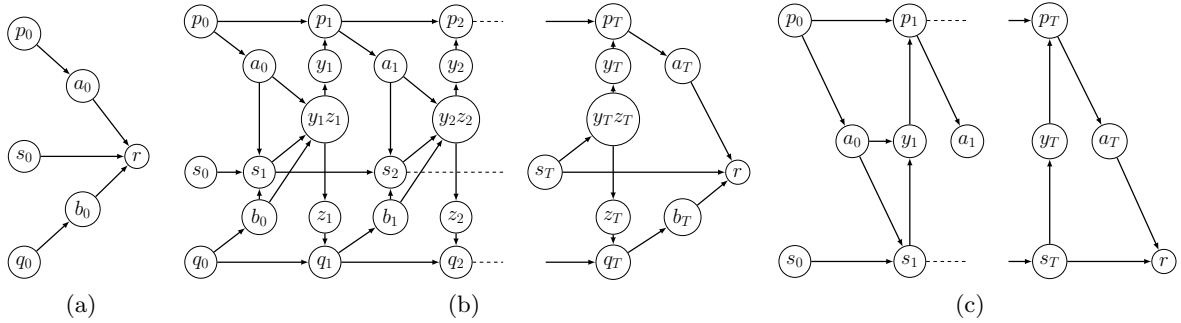

Figure 1: a) DEC-POMDP DBN for time step 0. b) for time step $T$. c) POMDP DBN for time step $T$

**Theorem 1.** *Let the CPT of binary rewards $r$ be such that $\hat{R}_{sab} \propto R_{sab}$ and the discounting time prior be set as $P(T) = \gamma^T(1-\gamma)$. Then, maximizing the likelihood $L^\theta = P(r=1;\theta)$ in the mixture of DBNs is equivalent to optimizing the DEC-POMDP policy. Furthermore, the joint-policy value relates linearly to the likelihood as $V^\theta = (R_{max} - R_{min})L^\theta/(1-\gamma) + \sum_T \gamma^T R_{min}$*

The proof is omitted as it is very similar to that of MDPs and POMDPs [19]. Before detailing the EM algorithm, we describe the DBN representation of DEC-POMDPs–the basis for any inference technique.

The DBN for any time step $T$ is shown in Fig. 1(b). Every node is a random variable with subscripts indicating time. $p_i$ denotes controller nodes for agent 1 and $q_i$ for agent 2. The remaining nodes represent the states, actions, and observations. There are four kinds of dependencies induced by the DEC-POMDP model that the DBN must represent:

- **State transitions:** State transitions as a result of the joint action of both agents and the previous state, shown by the DBN's middle layer.
- **Controller node transitions ($\lambda$):** These transitions depend on the last controller state and the most recent *individual* observation received. They are shown in the top and bottom layers.
- **Action probabilities ($\pi$):** The action taken at any time step $t$ depends on the current controller state. The links between controller nodes ($p_i$ or $q_i$) and action nodes ($a_i$ or $b_i$) model this.
- **Observation probabilities:** First, the probability of receiving joint observation $y_i z_i$ depends on the joint action of both agents and the domain state. This relationship is modeled by the DBN nodes labeled $y_i z_i$. Second, the individual observation each agent receives is a deterministic function of the joint observation. That is $P_{yy'z'} = P(y|y'z') = 1$ if $y = y'$ else 0. This is modeled by a link between $y_i z_i$ and the nodes $y_i$ and $z_i$.

To highlight the differences from a POMDP, Fig. 1(c) shows the DBN for a POMDP. The sheer scale of interactions present in a DEC-POMDP DBN become clear from this comparison, also highlighting the difficulty of solving DEC-POMDPs even approximately. In a POMDP, an agent receives the observation which is affected by the environment state, whereas in a DEC-POMDP agents only perceive the individual part of the joint observation $y_i z_i$. Such differences in the interaction structure make the E and M steps of a DEC-POMDP EM very different from that of a POMDP, despite sharing the same high-level principles.

## 4 EM algorithm for DEC-POMDPs

This section describes the EM algorithm [7] for maximizing the reward likelihood in the mixture of DBNs representing DEC-POMDPs. In the corresponding DBNs, only the binary reward is treated as observed ($r=1$); all other variables are latent. While maximizing the likelihood, EM yields the DEC-POMDP joint-policy parameters $\theta$. EM also possesses the desirable anytime characteristic as the likelihood (and the policy value which is proportional to the likelihood) is guaranteed to increase per iteration until convergence. We note that EM is not guaranteed to converge to the global optima. However, in the experiments we show that EM almost always achieves similar values as the state-of-the-art NLP based solver [1] and much better than DEC-BPI [4]. The main advantage of using EM lies in its ability to easily generalize to much richer representations than currently possible for DEC-POMDPs such as factored or hierarchical controllers, continuous state and action spaces. Another important advantage is the ability to generalize the solver to larger multi-agent systems with more than 2 agents.

The E step we derive next is generic as any probabilistic inference technique can be used.

## 4.1 E-step

In the E-step, for the fixed parameter $\theta$, forward messages $\alpha$ and backward messages $\beta$ are propagated. First, we define the following Markovian transitions on the $(p,q,s)$ state in the DBN of Fig. 1(b). These transitions are independent of the time $t$ due to the stationary joint policy. We also adopt the convention that for any random variable $v$, $v'$ refers to the next time slice and $\bar{v}$ refers to the previous time slice. For any group of variables $\mathbf{v}$, $P_t(\mathbf{v}, \mathbf{v}')$ refers to $P(\mathbf{v}_t = \mathbf{v}, \mathbf{v}_{t+1} = \mathbf{v}')$.
$$P(p', q', s'|p, q, s) =$$
$$\sum_{aby'z'} \lambda_{p'py'}\lambda_{q'qz'} P_{y'z'abs'} \pi_{ap}\pi_{bq} P_{s'sab} \quad (1)$$

$\alpha_t$ is defined as $P_t(p,q,s;\theta)$. It might appear that we need to propagate $\alpha$ messages for each DBN separately, but as pointed out in [19], only one sweep is required as the head of the DBN is shared among all the mixture components. That is, $\alpha_2$ is the same for all the T-step DBNs with $T \geq 2$. We will omit using $\theta$ as long as it is unambiguous.

$$\begin{aligned} \alpha_0(p,q,s) &= \nu_p \nu_q b_0(s) \\ \alpha_t(p', q', s') &= \sum_{p,q,s} P(p', q', s'|p,q,s) \alpha_{t-1}(p,q,s) \end{aligned}$$

Intuitively, $\alpha$ messages compute the probability of visiting a particular $(p,q,s)$ state in the DBN as per the current policy. The $\beta$ messages are similar to computing the value of starting the controllers in nodes $\langle p, q \rangle$ at state $s$ using dynamic programming. They are propagated backwards and are defined as $P_t(r=1|p,q,s)$. However, this particular definition would require separate inference for each DBN as for $T$ and $T'$ step DBN, $\beta_t$ will be different due to difference in the time-to-go ($T - t$ and $T' - t$). To circumvent this problem, $\beta$ messages are indexed backward in time as $\beta_\tau(p,q,s) = P_{T-\tau}(r=1|p,q,s)$ using the index $\tau$ such that $\tau = 0$ denotes the time slice $t = T$. Hence we get:

$$\begin{aligned} \beta_0(p,q,s) &= \sum_{ab} R_{sab} \pi_{ap} \pi_{bq} \\ \beta_\tau(p,q,s) &= \sum_{p',q',s'} \beta_{\tau-1}(p',q',s') P(p',q',s'|p,q,s) \end{aligned}$$

Based on the $\alpha$ and $\beta$ messages we also calculate two more quantities $\hat{\alpha}(p,q,s) = \sum_t P(T=t)\alpha(p,q,s)$ and $\hat{\beta}(p,q,s) = \sum_t P(T=t)\beta(p,q,s)$, which will be used in the M-step. The cut-off time for message propagation can either be fixed a priori or be more flexible based on the likelihood accumulation. If $\alpha$ messages are propagated for $t$-steps and $\beta$-messages for $\tau$ steps, then the likelihood for $T = t + \tau$ is given by

$$L^\theta_{t+\tau} = P(r=1|T=t+\tau;\theta) = \sum_{p,q,s} \alpha_t(p,q,s)\beta_\tau(p,q,s)$$

If both $\alpha$ and $\beta$ messages are propagated for $k$ steps and $L^\theta_{2k} \ll \sum_{T=0}^{2k-1} \gamma^T L^\theta_T$, then the message propagation can be stopped.

### 4.1.1 Complexity

Calculating the Markov transitions on the $(p,q,s)$ chain has complexity $O(N^4 S^2 A^2 Y^2)$, where $N$ is the maximum number of nodes for a controller. The message propagation has complexity $O(T_{max} N^4 S^2)$. Techniques to effectively reduce this complexity without sacrificing accuracy will be discussed later.

## 4.2 M-step

In the DBNs of Fig. 1(a,b) every variable is hidden except the reward variable. After each M-step, EM provides better estimates of these variables, improving the likelihood $L^\theta$ and hence the policy value. For details of EM, we refer to [7]. The parameters to estimate are $\langle \pi, \lambda, \nu \rangle$ for each agent. For a particular DBN for time $T$, let $\tilde{L} = (P, Q, A, B, S)$ denote the latent variables, where each variable denotes a sequence of length $T$. That is, $P = p_{0:T}$. EM maximizes the following expected complete log-likelihood for the DEC-POMDP DBN mixture. $\theta$ denotes the previous parameters and $\theta^\star$ denotes new parameters.

$$Q(\theta, \theta^\star) = \sum_T \sum_{\tilde{L}} P(r=1, \tilde{L}, T; \theta) \log P(r=1, \tilde{L}, T; \theta^\star)$$

In the rest of the section, all the derivations refer to the general DBN structure of the DEC-POMDP as in Fig. 1(b). The joint probability of all the variables is:

$$P(r=1, \tilde{L}, T; \theta) = P(T)[R_{sab}]_{t=T} \Big[\prod_{t=1}^{T} \pi_{ap}\pi_{bq} P_{s\bar{s}\bar{a}\bar{b}}$$
$$P_{yyz} P_{zyz} P_{yzs\bar{a}\bar{b}} \lambda_{p\bar{p}y} \lambda_{q\bar{q}z}\Big] \big[\pi_{ap}\pi_{bq}\nu_p\nu_q b_0(s)\big]_{t=0} \quad (2)$$

where brackets indicate the time slices, i.e., $[R_{sab}]_{t=T} = R(s_T, a_T, b_T)$. Taking the log, we get:

$$\log P(r=1, \tilde{L}, T) = \ldots + \sum_{t=0}^{T} \log \pi_{a_t p_t} + \sum_{t=0}^{T} \log \pi_{b_t q_t}$$
$$+ \sum_{t=1}^{T} \log \lambda_{p_t p_{t-1} y_t} + \sum_{t=1}^{T} \log \lambda_{q_t q_{t-1} z_t}$$
$$+ \log \nu_{p_0} + \log \nu_{q_0} \quad (3)$$

where the missing terms represents the quantities independent of $\theta$. As all the policy parameters $\langle \pi, \lambda, \nu \rangle$ get separated out for each agent in the log above, we first derive the action updates for an agent by substituting Eq. 3 in $Q(\theta, \theta^\star)$

### 4.2.1 Action updates

The update for action parameters $\pi^\star_{ap}$ for agent 1 can be derived by simplifying $Q(\theta, \theta^\star)$ as follows:

$$Q(\theta, \theta^\star) = \sum_{T=0}^{\infty} P(T) \sum_{t=0}^{T} \sum_{a,p} \left[P(r=1,a,p|T;\theta)\right]_t \log \pi^\star_{ap}$$

By breaking the above summation between $t=T$ and $t=0$ to $T-1$, we get

$$\sum_{T=0}^{\infty} P(T) \sum_{apqbs} R_{sab}\pi_{ap}\pi_{bq}\alpha_T(p,q,s) \log \pi^\star_{ap} + \sum_{T=0}^{\infty} P(T)$$

$$\sum_{t=0}^{T-1} \sum_{app'q's'} \beta_{T-t-1}(p',q',s') P_t(a,p,p',q',s') \log \pi^\star_{ap}$$

In the above equation, we marginalized the last time slice over the variables $(q,b,s)$. For the intermediate time slice $t$, we condition upon the variables $(p',q',s')$ in the next time slice $t+1$. We now use the definition of $\hat{\alpha}$ and move the summation over time $T$ inside for the last time slice and further marginalize over the remaining variables $(q,s)$ in the intermediate slice $t$:

$$= \sum_{a,p,q,b,s} R_{sab}\pi_{ap}\pi_{bq}\hat{\alpha}(p,q,s) \log \pi^\star_{ap} +$$

$$\sum_{T=0}^{\infty} P(T) \sum_{t=0}^{T-1} \sum_{ap} \log \pi^\star_{ap} \sum_{p'q's'sq} \beta_{T-t-1}(p',q',s')\pi_{ap}$$

$$P(p',q',s'|a,p,q,s)\alpha_t(p,q,s)$$

Upon further marginalizing over the joint observations $y'z'$ and simplifying we get:

$$= \sum_{ap} \pi_{ap} \log \pi^\star_{ap} \sum_{qs} \left[\sum_{b} R_{sab}\pi_{bq}\hat{\alpha}(p,q,s) + \sum_{p'q's'y'z'} \sum_{T=0}^{\infty} P(T) \sum_{t=0}^{T-1} \beta_{T-t-1}(p',q',s') P(s'|a,q,s) \right.$$

$$\left. \lambda_{p'py'}\lambda_{q'qz'}P(y'z'|a,q,s')\alpha_t(p,q,s)\right]$$

We resolve the above time summation, as in [19], based on the fact that $\sum_{T=0}^{\infty}\sum_{t=0}^{T-1} f(T-t-1)g(t)$ can be rewritten as $\sum_{t=0}^{\infty}\sum_{T=t+1}^{\infty} f(T-t-1)g(t)$ and then setting $\tau = T-t-1$ to get $\sum_{t=0}^{\infty} g(t) \sum_{\tau=0}^{\infty} f(\tau)$. Finally we get:

$$= \sum_{ap} \pi_{ap} \log \pi^\star_{ap} \sum_{qs} \hat{\alpha}(p,q,s) \left[\sum_b R_{sab}\pi_{bq} + \frac{\gamma}{1-\gamma}\right.$$

$$\left.\sum_{p'q's'y'z'} \hat{\beta}(p',q',s')\lambda_{p'py'}\lambda_{q'qz'}P(s'|a,q,s)P(y'z'|a,q,s')\right]$$

The product $P(s'|a,q,s)P(y'z'|a,q,s')$ can be further simplified by marginalizing out over actions $b$ of agent 2 as follows:

$$= \sum_{ap} \pi_{ap} \log \pi^\star_{ap} \sum_{qs} \hat{\alpha}(p,q,s)\left[\sum_b R_{sab}\pi_{bq} + \frac{\gamma}{1-\gamma}\right.$$

$$\left.\sum_{p'q's'y'z'} \hat{\beta}(p',q',s')\lambda_{p'py'}\lambda_{q'qz'} \sum_b P_{y'z's'ab}\pi_{bq}P_{s'sab}\right]$$

The above expression is maximized by setting the parameter $\pi^\star_{ap}$ to be:

$$\pi^\star_{ap} = \frac{\pi_{ap}}{C_p} \sum_{qs} \hat{\alpha}(p,q,s)\left[\sum_b R_{sab}\pi_{bq} + \frac{\gamma}{1-\gamma}\right.$$

$$\left.\sum_{p'q's'y'z'} \hat{\beta}(p',q',s')\lambda_{p'py'}\lambda_{q'qz'} \sum_b P_{y'z's'ab}\pi_{bq}P_{s'sab}\right] \quad (4)$$

where $C_p$ is a normalization constant. The action parameters $\pi^\star_{bq}$ of the other agent can be found similarly by the analogue of the previous equation.

### 4.2.2 Controller node transition updates

The update for controller node transition parameters $\lambda_{p\bar{p}y}$ for agent 1 can be found by maximizing $Q(\theta,\theta^\star)$ w.r.t $\lambda^\star_{p\bar{p}y}$ as follows.

$$Q(\theta,\theta^\star) = \sum_{T=0}^{\infty} P(T) \sum_{t=1}^{T} \sum_{p\bar{p}y} \left[P(r=1,p,\bar{p},y|T;\theta)\right]_t \log \lambda^\star_{p\bar{p}y}$$

By marginalizing over the variables $(q,s)$ for the current time slice $t$, we get

$$= \sum_{T=0}^{\infty} P(T) \sum_{t=1}^{T} \sum_{p\bar{p}ysq} \log \lambda^\star_{p\bar{p}y}\beta_{T-t}(p,q,s)P_t(p,\bar{p},y,s,q|T;\theta)$$

By further marginalizing over the variables $(\bar{s},\bar{q})$ for the previous time slice of $t$ and over the observations $z$ of the other agent, we get

$$= \sum_{p\bar{p}y} \lambda_{p\bar{p}y} \log \lambda^\star_{p\bar{p}y} \sum_{T=0}^{\infty} P(T) \sum_{t=1}^{T} \sum_{sq\bar{s}\bar{q}z} \beta_{T-t}(p,q,s)\lambda_{q\bar{q}z}$$

$$P(yz|\bar{p},\bar{q},s)P(s|\bar{p},\bar{q},\bar{s})\alpha_{t-1}(\bar{p},\bar{q},\bar{s})$$

The above equation can be further simplified by marginalizing the product $P(yz|\bar{p},\bar{q},s)P(s|\bar{p},\bar{q},\bar{s})$ over actions $a$ and $b$ of both the agents as follows:

$$= \sum_{p\bar{p}y} \lambda_{p\bar{p}y} \log \lambda^\star_{p\bar{p}y} \sum_{T=0}^{\infty} P(T) \sum_{t=1}^{T} \sum_{sq\bar{s}\bar{q}z} \beta_{T-t}(p,q,s)\lambda_{q\bar{q}z}$$

$$\alpha_{t-1}(\bar{p},\bar{q},\bar{s}) \sum_{ab} P_{yzsab}P_{s\bar{s}ab}\pi_{a\bar{p}}\pi_{b\bar{q}}$$

Upon resolving the time summation as before, we get the final M-step estimate:

$$\lambda^\star_{p\bar{p}y} = \frac{\lambda_{p\bar{p}y}}{C_{\bar{p}y}} \sum_{sq\bar{s}\bar{q}z} \hat{\alpha}(\bar{p}, \bar{q}, \bar{s}) \hat{\beta}(p, q, s) \lambda_{q\bar{q}z}$$
$$\sum_{ab} P_{yzsab} P_{s\bar{s}ab} \pi_{a\bar{p}} \pi_{b\bar{q}} \qquad (5)$$

The parameters $\lambda^\star_{q\bar{q}z}$ for the other agent can be found in an analogous way.

### 4.2.3 Initial node distribution

The initial node distribution $\nu$ for controller nodes of agent 1 and 2 can be updated as follows. We do not show the complete derivation as it is similar to that of the other parameters.

$$\nu^\star_p = \frac{\nu_p}{C_p} \sum_{qs} \hat{\beta}(p,q,s) \nu_q P_s b_0(s) \qquad (6)$$

### 4.2.4 Complexity and implementation issues

The complexity of updating all action parameters is $O(N^4 S^2 AY^2)$. Updating node transitions requires $O(N^4 S^2 Y^2 + N^2 S^2 Y^2 A^2)$. This is relatively high when compared to the POMDP updates requiring $O(N^2 S^2 AY)$ mainly due to the scale of the interactions present in DEC-POMDPs.

In our experimental settings, we observed that having a relatively small sized controller ($N \leq 5$) suffices to yield good quality solutions. The main contributor to the complexity is the factor $S^2$ as we experimented with large domains having nearly 250 states. The good news is that the structure of the E and M-step equations provides a way to effectively reduce this complexity by significant factor without sacrificing accuracy. For a given state $s$, joint action $\langle a, b \rangle$ and joint observation $\langle y, z \rangle$, the possible next states can be calculated as follows: $succ(s, a, b, y, z) = \{s' | P(s'|s, a, b) P(y, z|s', a, b) > 0\}$. For most of the problems, the size of this set is typically a constant $k < 10$. Such simple reachability analysis and other techniques could speed up the EM algorithm by more than an order of magnitude for large problems. The effective complexity reduces to $O(N^4 SAY^2 k)$ for the action updates and $O(N^4 SY^2 k + N^2 SY^2 A^2 k)$ for node transitions. Other enhancements of the EM implementation are discussed in Section 6.

## 5 Experiments

We experimented with several standard 2-agent DEC-POMDP benchmarks with discount factor 0.9. Complete details of these problems can be found in [1, 4].

| Size | DEC-BPI | NLP | EM | DEC-BPI | EM |
|---|---|---|---|---|---|
| 1 | 4.687 | 9.1 | 9.05 | < 1s | < 1s |
| 2 | 4.068 | 9.1 | 9.05 | < 1s | < 1s |
| 3 | 8.637 | 9.1 | 9.05 | 2s | 1.7s |
| 4 | 7.857 | 9.1 | 9.05 | 5s | 4.62s |

Table 1: Broadcast channel: Policy value, execution time

We compare our approach with the decentralized bounded policy iteration (DEC-BPI) algorithm [4] and a non-convex optimization solver (NLP) [1]. The DEC-BPI algorithm iteratively improves the parameters of a node using a linear program while keeping the other nodes' parameters fixed. The NLP approach recasts the policy optimization problem as a non-linear program and uses an off-the-shelf solver, Snopt [9], to obtain a solution. We implemented the EM algorithm in JAVA. All our experiments were on a Mac with 4GB RAM and 2.4GHz CPU. Each data point is an average of 10 runs with random initial controller parameters. In terms of solution quality, EM is always better than DEC-BPI and it achieves similar or higher solution quality than NLP. We note that our current implementation is mainly a proof-of-concept; we have not yet implemented several enhancements (discussed later) that could improve the performance of the EM approach. In contrast, the NLP solver [9] is an optimized package and therefore for larger problems is currently faster than the EM approach. The fact that a crude implementation of the EM approach works so well is very encouraging.

Table 1 shows results for the broadcast channel problem, which has 4 states, 2 actions per agent and 5 observations. This is a networking problem where agents must decide whether or not to send a message on a shared channel and must avoid collision to get a reward. We tested with different controller sizes. On this problem, all the algorithms compare reasonably well, with EM being better than DEC-BPI and very close in value to NLP. The time for NLP is also $\approx 1s$.

Fig. 2(a) compares the solution quality of the EM approach against DEC-BPI and NLP for varying controller sizes on the recycling robots problem. In this problem, two robots have the task of picking up cans in an office building. They can search for a small can, a big can or recharge the battery. The large item is only retrievable by the joint action of the two robots. Their goal is to coordinate their actions to maximize the joint reward. EM(2) and NLP(2) show the results with controller size 2 for both agents in Fig. 2(a). For this problem, EM works much better than both DEC-BPI and the NLP approach. EM achieves a value of $\approx 62$ for all controller sizes, providing nearly 12% improvement over DEC-BPI ($= 55$) and 20% improvement over NLP ($= 51$). Fig. 2(b) shows the time comparisons for EM with different controller sizes. Both the NLP and DEC-BPI take nearly 1s to converge. EM with controller size 2 has comparable performance, but as expected, EM with 4-node controllers takes longer as the complexity of EM is proportional to $O(N^4)$.

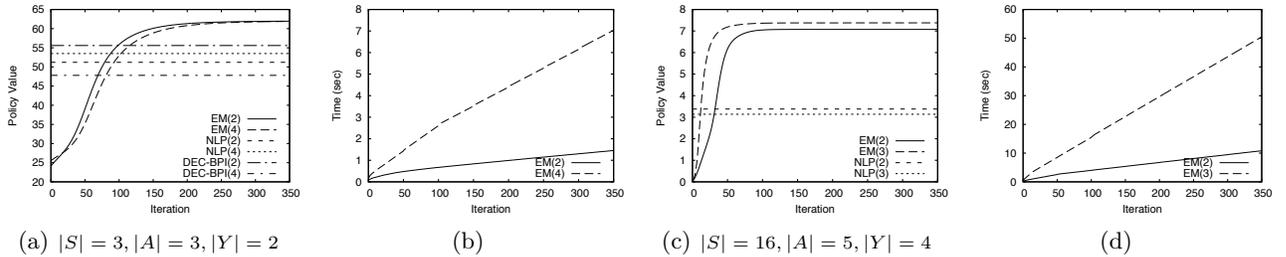

(a) $|S|=3, |A|=3, |Y|=2$  (b)  (c) $|S|=16, |A|=5, |Y|=4$  (d)

Figure 2: Solution quality and runtime for recycling robots (a) & (b) and meeting on a grid (c) & (d)

Fig. 2(c) compares the solution quality of EM on the meeting on a grid problem. In this problem, agents start diagonally across in a $2 \times 2$ grid and their goal is to take actions such that they meet each other (i.e., share the same square) as much as possible. As the figure shows, EM provides much better solution quality than the NLP approach. EM achieves a value of $\approx 7$, which nearly *doubles* the solution quality achieved by NLP ($= 3.3$). DEC-BPI results are not plotted as it performs much worse and achieves a solution quality of 0, essentially unable to improve the policy at all even for large controllers. Both DEC-BPI and NLP take around $1s$ to converge. Fig. 2(d) shows the time comparison for EM versions. EM with 2-node controllers is very fast and takes $< 1s$ to converge (50 iterations). Also note that in both the cases, EM could run with much larger controller sizes ($\approx 10$), but the increase in size did not provide tangible improvement in solution quality.

Fig. 3 shows the results for the multi-agent tiger problem, involving two doors with a tiger behind one door and a treasure behind the other. Agents should coordinate to open the door leading to the treasure [1]. Fig. 3(a) shows the quality comparisons. EM does not perform well in this case; even after increasing the controller size, it achieves a value of $-19$. NLP works better with large controller sizes. However, this experiment presents an interesting insight into the workings of EM as related to the scaling of the rewards. Recalling the relation between the likelihood and the policy value from Theorem 1, the equation for this problem is: $V^\theta = 1210L^\theta - 1004.5$. For EM to achieve the same solution as the best NLP setting ($= -3$), the likelihood should be .827. Fig. 3(b) shows that the likelihood EM converges to is .813. Therefore, from EM's perspective, it is finding a really good solution. Thus, the scaling of rewards has a significant impact (in this case, adverse) on the policy value. This is a potential drawback of the EM approach, which applies to other Markovian planning problems too when using the technique of [19]. Incidently, DEC-BPI performs much worse on this problem and gets a quality of $-77$.

Fig. 4 shows the results for the two largest DEC-POMDP domains–box pushing and Mars rovers. In the box pushing domain, agents need to coordinate and push boxes into a goal area. In the Mars rovers domain, agents need to coordinate their actions to perform experiments at multiple sites. Fig. 4(a) shows that EM performs much better than DEC-BPI for every controller size. For controller size 2, EM achieves better quality than NLP with comparable runtime (Fig. 4(b), 500 iterations). However, for the larger controller size ($= 3$), it achieves slightly lower quality than NLP. For the largest Mars rovers domain (Fig. 4(c)), EM achieves better solution quality ($= 9.9$) than NLP ($= 8.1$). However, EM also takes many more iterations to converge than for previous problems and hence, requires more time than NLP. EM is also much better than DEC-BPI, which achieves a quality of $-1.18$ and takes even longer to converge (Fig. 4(d)).

## 6 Conclusion and future work

We present a new approach to solve DEC-POMDPs using inference in a mixture of DBNs. Even a simple implementation of the approach provides good results. Extensive experiments show that EM is always better than DEC-BPI and compares favorably with the state-of-the-art NLP solver. The experiments also highlight two potential drawbacks of the EM approach: the adverse effect of reward scaling on solution quality and slow convergence rate for large problems. We are currently addressing the runtime issue by parallelizing the algorithm. For example, $\alpha$ and $\beta$ can be propagated in parallel. Even updating each node's parameters can be done in parallel for each iteration. Furthermore, the structure of EM's update equations is very amenable to Google's Map-Reduce paradigm [6], allowing each parameter to be computed by a cluster of machines in parallel using Map-Reduce. Such scalable techniques will certainly make our approach many times faster than the current *serial* implementation. We are also investigating how a different scaling of rewards affects the convergence properties of EM.

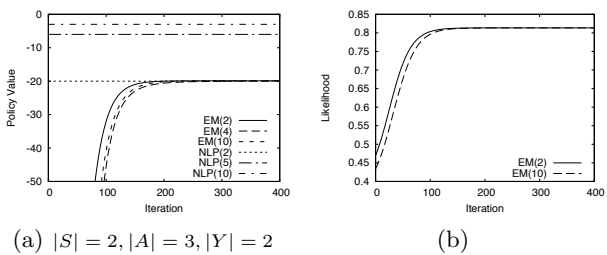

(a) $|S|=2, |A|=3, |Y|=2$  (b)

Figure 3: Solution quality (a) and likelihood (b) for "tiger"

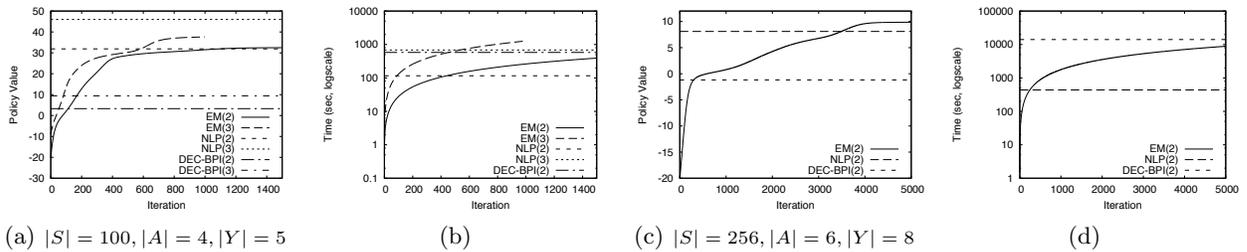

(a) $|S| = 100, |A| = 4, |Y| = 5$   (b)   (c) $|S| = 256, |A| = 6, |Y| = 8$   (d)

Figure 4: Solution quality and runtime for box pushing (a) & (b) and Mars rovers (c) & (d)

The main benefit of the EM approach is that it opens up the possibility of using powerful probabilistic inference techniques to solve decentralized planning problems. Using a graphical DBN structure, EM can easily generalize to richer representations such as factored or hierarchical controllers, or continuous state and action spaces. Unlike the existing techniques, EM can easily extend to larger multi-agent systems with more than 2 agents. The ND-POMDP model [13] is a class of DEC-POMDPs specifically designed to support large multi-agent systems. It makes some restrictive yet realistic assumptions such as locality of interaction among agents, and transition and observation independence. EM can naturally exploit such independence structure in the DBN and scale to larger multi-agent systems, something that current infinite-horizon algorithms fail to achieve. Hence the approach we introduce offers great promise to overcome the shortcomings of the prevailing approaches to multi-agent planning.

## Acknowledgments

Support for this work was provided in part by the National Science Foundation Grant IIS-0812149 and by the Air Force Office of Scientific Research Grant FA9550-08-1-0181.

## References

[1] C. Amato, D. S. Bernstein, and S. Zilberstein. Optimizing fixed-size stochastic controllers for POMDPs and decentralized POMDPs. *JAAMAS*, 2009.

[2] H. Attias. Planning by probabilistic inference. In *Workshop on AISTATS*, 2003.

[3] R. Becker, S. Zilberstein, V. Lesser, and C. V. Goldman. Solving transition independent decentralized markov decision processes. *JAIR*, 22:423–455, 2004.

[4] D. S. Bernstein, C. Amato, E. A. Hansen, and S. Zilberstein. Policy iteration for decentralized control of Markov decision processes. *JAIR*, 34:89–132, 2009.

[5] D. S. Bernstein, R. Givan, N. Immerman, and S. Zilberstein. The complexity of decentralized control of Markov decision processes. *J. MOR*, 27:819–840, 2002.

[6] J. Dean and S. Ghemawat. MapReduce: a flexible data processing tool. *CACM*, 53(1):72–77, 2010.

[7] A. P. Dempster, N. M. Laird, and D. B. Rubin. Maximum likelihood from incomplete data via the EM algorithm. *Journal of the Royal Statistical society, Series B*, 39(1):1–38, 1977.

[8] J. S. Dibangoye, A.-I. Mouaddib, and B. Chaib-draa. Point-based incremental pruning heuristic for solving finite-horizon DEC-POMDPs. In *AAMAS*, pages 569–576, 2009.

[9] P. E. Gill, W. Murray, and M. A. Saunders. SNOPT: An SQP algorithm for large-scale constrained optimization. *SIOPT*, 12(4):979–1006, 2002.

[10] M. Hoffman, H. Kueck, N. de Freitas, and A. Doucet. New inference strategies for solving Markov decision processes using reversible jump MCMC. In *UAI*, 2009.

[11] A. Kumar and S. Zilberstein. Point based backup for decentralized POMDPs: Complexity and new algorithms. In *AAMAS*, pages 1315 – 1322, 2010.

[12] M. Mundhenk, J. Goldsmith, C. Lusena, and E. Allender. Complexity of finite-horizon Markov decision process problems. *J. ACM*, 47(4):681–720, 2000.

[13] R. Nair, P. Varakantham, M. Tambe, and M. Yokoo. Networked distributed POMDPs: A synthesis of distributed constraint optimization and POMDPs. In *AAAI*, pages 133–139, 2005.

[14] F. A. Oliehoek, M. T. J. Spaan, and N. A. Vlassis. Optimal and approximate Q-value functions for decentralized POMDPs. *JAIR*, 32:289–353, 2008.

[15] J. Pineau, G. Gordon, and S. Thrun. Anytime point-based approximations for large POMDPs. *JAIR*, 27:335–380, 2006.

[16] S. Seuken and S. Zilberstein. Memory-bounded dynamic programming for DEC-POMDPs. In *IJCAI*, pages 2009–2015, 2007.

[17] T. Smith and R. Simmons. Heuristic search value iteration for POMDPs. In *UAI*, pages 520–527, 2004.

[18] M. Toussaint, L. Charlin, and P. Poupart. Hierarchical POMDP controller optimization by likelihood maximization. In *UAI*, pages 562–570, 2008.

[19] M. Toussaint, S. Harmeling, and A. Storkey. Probabilistic inference for solving (PO)MDPs. Technical Report EDIINF-RR-0934, University of Edinburgh, School of Informatics, 2006.

[20] M. Toussaint and A. J. Storkey. Probabilistic inference for solving discrete and continuous state markov decision processes. In *ICML*, pages 945–952, 2006.